\documentclass[letterpaper, 10 pt, journal]{ieeeconf}

\IEEEoverridecommandlockouts                              

\usepackage{etoolbox}
\makeatletter
\patchcmd{\@makecaption}
  {\scshape}
  {}
  {}
  {}
\makeatletter
\patchcmd{\@makecaption}
  {\\}
  {.\ }
  {}
  {}
\makeatother

\usepackage{graphics} 
\usepackage{epsfig} 
\usepackage{mathptmx} 
\usepackage{times} 
\usepackage{amsmath} 
\usepackage{amssymb}  
\usepackage{algorithm}
\usepackage{algpseudocode}
\usepackage{cite}
\usepackage{hyperref}
\usepackage{booktabs}
\usepackage{caption}
\usepackage{makecell} 
\usepackage{multirow}
\usepackage{adjustbox}
\usepackage{changepage}
\usepackage{xcolor}
\usepackage{wrapfig}
\usepackage{subcaption}

\usepackage{geometry}
\graphicspath{{Figures/}}

\hypersetup{
  colorlinks   = true,    
  urlcolor     = blue,    
  linkcolor    = blue,    
  citecolor    = blue      
}

\title{\LARGE \bf Improving the Generalization of Unseen Crowd Behaviors for Reinforcement Learning based Local Motion Planners}

\author{Wen Zheng Terence Ng$^{1,2}$, Jianda Chen$^{1}$, Sinno Jialin Pan$^{3}$, Tianwei Zhang$^{1}$ 
\\
{\small$^1$Nanyang Technological University, $^2$Continental Automotive Singapore, $^3$The Chinese University of Hong Kong}\\
{\tt\small\{ngwe0099, jianda001\}@e.ntu.edu.sg, sinnopan@cuhk.edu.hk, tianwei.zhang@ntu.edu.sg }\\ 
}      

\begin{document}

\maketitle
\thispagestyle{empty}
\pagestyle{empty}

\begin{abstract}

Deploying a safe mobile robot policy in scenarios with human pedestrians is challenging due to their unpredictable movements. Current Reinforcement Learning-based motion planners rely on a single policy to simulate pedestrian movements and could suffer from the over-fitting issue. Alternatively, framing the collision avoidance problem as a multi-agent framework, where agents generate dynamic movements while learning to reach their goals, can lead to conflicts with human pedestrians due to their homogeneity.

To tackle this problem, we introduce an efficient method that enhances agent diversity within a single policy by maximizing an information-theoretic objective. This diversity enriches each agent's experiences, improving its adaptability to unseen crowd behaviors.
In assessing an agent's robustness against unseen crowds, we propose diverse scenarios inspired by pedestrian crowd behaviors. Our behavior-conditioned policies outperform existing works in these challenging scenes, reducing potential collisions without additional time or travel.

\end{abstract}

\section{Introduction}

Mobile robots are increasingly used in various applications ranging from industrial automation, service delivery, to agriculture applications \cite{ben2018robots}. The ability of these robots to maneuver and navigate in complex and dynamic environments is crucial for their successful deployment. One key aspect of mobile robot navigation is \textit{local motion planning}, which aims to find a feasible and safe path for the robot to follow in its immediate vicinity. This task is particularly challenging, as it needs to ensure safe, efficient, and smooth robot movements in the presence of  dynamic obstacles (i.e., pedestrians) in the environment. To address this issue, Reinforcement Learning (RL) has been introduced to achieve local motion planning, which exhibits the high ability to handle more complex scenarios and increased levels of uncertainty \cite{tai2017virtual,chen2017decentralized,chen2019crowd,long2018towards,pfeiffer2018reinforced}.

For RL-based methods, the environment is crucial as it shapes the agent's understanding of the environment to train an optimal policy. 
Particularly, the scenes in the environment should fully reflect the inherent diversity and unpredictability of pedestrians' movements. For example, on the footpaths, human pedestrians may walk at different speeds or behave differently depending on their social norms. If their behaviors are not modeled comprehensively, it is challenging for the robot agent to learn a robust policy which works well against diverse and unseen crowd behaviors.

Various approaches have been proposed to generate pedestrian movements for training RL-based local motion planning policies, which can be classified into two categories. Unfortunately, both suffer limitations. 
The first category involves single-agent approaches. One simple approach is to assign pedestrians' waypoints from a dataset \cite{fan2020learning}, which doesn't enable interactions between robots and pedestrians, limiting their influence on each other. To address this, some works manually design pedestrians' behaviors based on crowd density \cite{fan2019getting, liang2021crowd}, or use fixed non-learning-based algorithms to control pedestrians \cite{jin2020mapless,zhou2021r,brito2021go,chen2019crowd}. However, these approaches may lead to agents overfitting due to limited diversity in generated pedestrian movements.

\begin{figure}[t]
    \centering
    \includegraphics[width=0.85\linewidth]{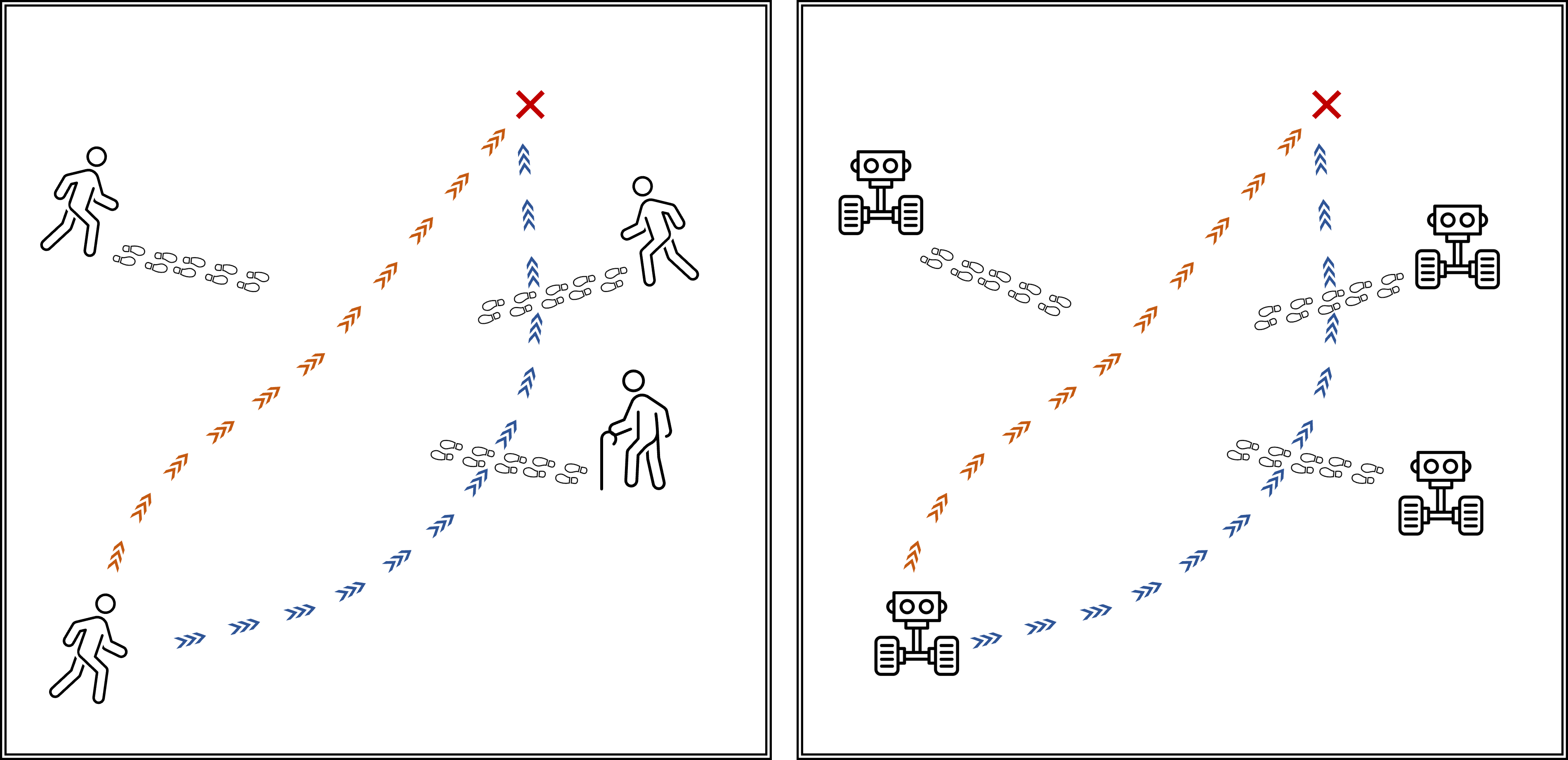}
    \caption{\textbf{A human may take \textcolor{black}{diverse} strategies to reach the same predefined goal (left). \textcolor{black}{We propose a behavior-conditioned policy to integrate such diversity into the robot agent (right). This diversity enriches the agent with a more varied range of experiences when learning in a multi-agent framework, and improves its ability to generalize in unseen crowd behaviors.}}}
    \label{fig:high}
\vspace{-20pt}
\end{figure}

The second approach frames local motion planning as a decentralized multi-agent collision avoidance task \cite{chen2017socially,chen2017decentralized,long2018towards}. Each agent is linked to a single policy to reach its goal, learning to avoid collisions and adapting dynamically during training. 
This strategy offers two advantages: (1) it doesn't require explicit specification of each pedestrian's behavior to avoid bias; (2) it's highly sample-efficient due to bootstrapping, as every agent's trajectory can update the same model. However, these solutions face practical challenges when applied to scenarios with diverse or unforeseen dynamics.
The learned policy assumes homogeneous behaviors among multiple robots upon deployment, posing a challenge for navigating in scenarios with varied crowd behaviors where such assumptions don't hold.

In this paper, we propose a novel sample-efficient multi-agent framework to enhance behavior diversity among agents. Diverse movements induced by different agents enrich experiences and enhance robustness to unpredictable behaviors in unseen or challenging scenarios.
Our framework introduces the concept of behaviors conditioned on a policy. These behaviors, randomly sampled as token embeddings by each agent at the start of each training episode, incentivize agent diversity. We assign intrinsic rewards for agents to take varied actions for every state conditioned on the sampled behavior. These rewards are based on a discriminator capable of identifying behavior from a state-action pair.
With this approach, we can train robust RL policies for local motion planning in highly complex environments.

We perform comprehensive simulation experiments to evaluate the robustness of our framework on a diverse set of unseen pedestrians' behaviors.
Simulation results show that the proposed behavior-conditioned policy is more robust while \textcolor{black}{having the same number of updates}. 

\section{Preliminaries}

\subsection{RL-based Local Motion Planning}
Local motion planning is a sequential decision-making task that can be formulated as a Markov Decision Process (MDP),  defined by a tuple $M=\langle S, A, P, R, \gamma\rangle$. Here, $S$ is the state space, $A$ is the action space, $P$ is the state-transition model, $R$ is the reward function, and $\gamma$ is a discount factor.

A general form of the states is $s=[s_{env}, s_{robot}, s_{goal}]$, where $s_{env}$, $s_{robot}$ and $s_{goal}$ contain information about the observed environment, robot and goal respectively. 
Similar to \cite{long2018towards}, we choose a realistic representation that uses distance readings from a 2D laser range finder to sense the environment $s_{env}$. We consider the sensor noise and obstacle occlusions, and make no assumptions about the shape, size, and number of obstacles, which are more closely aligned with the real world. $s_{robot}$ reveals the state of the robot, usually the velocities and optionally the position. $s_{goal}$ is represented by either the relative or absolute goal position. 
The action space $A$ is the set of permissible velocities in either the continuous or discrete space. The reward $R$ can be normally represented as follows: 
\begin{equation*}
r_t = \begin{cases}
r_{goal} & \text { if }\left\|\mathbf{p}_t-\mathbf{g}\right\|< d_{col}, \\ 
r_{col} & \text { else if collision }, \\ 
r_{step} \cdot\left(\left\|\mathbf{p}_{t-1}-\mathbf{g}\right\|-\left\|\mathbf{p}_t-\mathbf{g}\right\|\right) & \text { otherwise, }\end{cases}
\label{eq:1}
\end{equation*}
where $r_{goal}$ is the reward for reaching the desired goal, $r_{col}$ is the penalty for collision, $r_{step}$ is the dense reward for getting closer to the goal, $d_{col}$ is the distance threshold for reaching the goal, $\mathbf{p}$ and $\mathbf{g}$ are positions of the robot and goal.

In this MDP, we aim to use RL to find a policy $\pi_\theta$ parametrized by $\theta$, which maps states to actions and maximizes the expected sum of discounted rewards,
$
J(\boldsymbol{\theta})=\mathbb{E}\left[\sum_{t=0}^T \gamma^t \cdot r_t\right],
$
where $T$ is the length of an episode.

\subsection{Pedestrian Modelling}
In local motion planning, the movement of dynamic obstacles, often represented by human pedestrians, is a crucial environmental factor. Each pedestrian's behavior significantly influences the environment and how the robot agent learns within the RL framework. Existing approaches to modeling pedestrians can be classified into two categories. 

\noindent\textbf{Single-agent approaches.}
In these approaches, only a single robot agent learns to navigate within the crowd, while pedestrians are typically modeled using non-learning-based algorithms. 
Some examples of the non-learning-based algorithms include Velocity Obstacles \cite{fiorini1998motion}, Social Forces \cite{jin2020mapless,zhou2021r,brito2021go,chen2019crowd} and physics inspired movements \cite{wang2018learning}. One common drawback to these approaches is that the RL-agent might overfit to the chosen pedestrian behaviors during training.

\noindent\textbf{Multi-agent approaches.}
The main idea of the multi-agent framework \cite{chen2017decentralized, long2018towards,everett2018motion,tan2020deepmnavigate,semnani2020multi,cui2021learning,fan2020distributed, chen2017socially} is to control multiple agents with a single policy. In this setup, agents must learn to avoid each other to reach their goals, leading to the emergence of \textcolor{black}{dynamic movements} during the course of training. One consequence under this framework is that the agents converge to \textcolor{black}{homogeneous behaviors as all agents boot-strap to a single policy}.

\subsection{Behavior Diversity in RL}
\label{sec:behavior-diversity}
To address the above homogeneity concern, various approaches were proposed to increase agent behavior diversity.
For single-agent scenarios, one popular solution \cite{haarnoja2017reinforcement} is to maximize the entropy of the policy in addition to the reward, to learn different behaviors to achieve the goal. 
Eysenbac et al. \cite{eysenbach2018diversity} introduced DIYAN to increase the diversity of agent behaviors by maximizing the mutual information between skills and states, resulting in better state exploration.

In the multi-agent scenario, some work increases the behavior diversity of multiple agents in the Centralized Training with Decentralized Execution (CTDE) framework \cite{lowe2017multi,foerster2016learning,rashid2020monotonic}. When agents are assigned different tasks, each agent gets a distinct policy respectively, which shares a common critic network. With multiple policies, several ideas have been proposed to generate diverse behaviors among agents \cite{mckee2020social, li2021celebrating, lee2020learning}. However, this comes at the expense of sample efficiency since each agent only updates its own policy instead of a unified policy.

\section{Approach}
We present our approach to learning robust agents through behavior diversity. Instead of using multiple policies to create diversity as in CTDE, we opt for a more sample-efficient method by using only a single policy.
We first formulate the agent behaviors, and how they can be used to generate diversity among agents (Section \ref{sec:behavior-formulation}). Then we explain how the behaviors and diversity can be integrated together within a single policy (Section \ref{sec:behavior-policy}). Finally, we describe how to train a behavior-conditioned policy in an end-to-end manner with all the integrated components (Section \ref{sec:training}). 

\subsection{Agent Behavior} 
\label{sec:behavior-formulation}
In Figure \ref{fig:high}, people's approaches to walking towards a goal can vary: some prioritize speed with a longer path, while others choose a shorter route at a slower pace. When avoiding moving obstacles, some turn left, while others turn right. Although individuals may have unique behaviors, there can be similar patterns. We formalize this with discrete behavior tokens $z \in [0,1, \dots, M-1]$, where $M$ is the total number of distinct behaviors. Each token represents a distinct pedestrian behavior, and different agents may share the same token.

To foster diversity amongst different agents, our goal is to assign different behavior tokens to different agents to exhibit distinct behaviors. In other words, for every state $s$, agents should perform different actions $a$ depending on the assigned $z$. 
More formally, this idea can be formalised using information theory by maximizing the mutual information $I( (S,A);Z)$, where $(S,A)$ is the joint distribution of $S$ and $A$. $Z \sim p(z)$, $S$, and $A$ represent the random variables for behavior, state, and action respectively. Additionally, the diverse actions performed for different $z$ should arise for every state instead of exploiting only certain states. For this, we minimize $I(S,Z)$ as a regularizer. In sum, we maximize 
\begin{equation}
\begin{aligned}
\mathcal{F}(\theta) & \triangleq I( (S,A) ; Z) - I( S ; Z) \\
& =({H}[Z]-{H}[Z \mid S,A]) - ({H}[Z]-{H}[Z \mid S])\\
& = - {H}[Z \mid S,A] + {H}[Z \mid S],
\end{aligned}
\label{eq:reg}
\end{equation}
where $H$ is the Shannon entropy. The first term implies it is easy to infer the behavior $z$ given any $(s,a)$. This makes sense intuitively as it means the agents are distinguishable due to their diverse behaviors and not behaving in a homogeneous way. The second term implies that the agents' behavior should not be distinguishable exclusively given $s$. 
It is intractable to compute $p(z|s)$  and $ p(z|(s,a))$ by integrating all states, actions, and skills. So we approximate the posteriors with learned discriminators $q_\phi(z | s)$ and $q_\psi(z | (s,a))$. We instead optimize the variational lower bound derived using Jensen’s Inequality \cite{barber2004algorithm}:
$$
\begin{aligned}
\mathcal{F}(\theta) & = - H[Z \mid S,A] + H[Z \mid S]  \\
& = \mathbb{E}_{z \sim p(z), s \sim \pi(z)}[\log p(z \mid s)] \\
& \quad - \mathbb{E}_{z \sim p(z), s \sim \pi(z), a \sim \pi(s,z)}[\log p(z \mid s,a)] \\
& \geq \mathbb{E}_{z \sim p(z), s \sim \pi(z)}[\log q_\phi(z \mid s)] \\
& \quad - \mathbb{E}_{z \sim p(z), s \sim \pi(z), a \sim \pi(s,z)}[\log q_\psi(z \mid s,a)] \\
& \triangleq \mathbb{G}(\theta),
\end{aligned}
$$
where $s\sim\pi(z)$ means to first sample the action $a$ from $\pi$ followed by sampling the environment to get the state $s$. 
It is non-trivial to directly optimize $\theta$ via maximizing the lower bound $\mathbb{G}(\theta)$ since $s\sim\pi(z)$ has to be sampled through a non-differentiable simulator. Below we introduce how to optimize $\theta$ using an intrinsic reward alongside the RL objective.

\subsection{Behavior-Conditioned Policy}
\label{sec:behavior-policy}
First, we incorporate the idea of behaviors into our policy where we condition our policy on the agent's behaviors. 
Each agent, $i$, sample their actions from a shared behavior-conditioned policy as 
$
a \sim \pi_\theta ( \cdot, s_t^i | z^i ),
$
for behavior token ID $z^i$ at timestep $t$. 
Each behavior token maps to an embedding in the policy network, enabling the policy to generate distinct behaviors for agents. To maximize such diversity, we introduce an intrinsic \textcolor{black}{pseudo-reward} $r_{int}$ motivated from maximizing $\mathbb{G}(\theta)$ derived previously:
\begin{equation}
r^{int}_t = log [q_{\psi_{sa}}(z \mid s_t,a_t)] - log[q_{\psi_{s}}(z \mid s_t)].
\label{eq:2} 
\end{equation}
 
\begin{figure}[t]
    \centering
    \includegraphics[width=\linewidth]{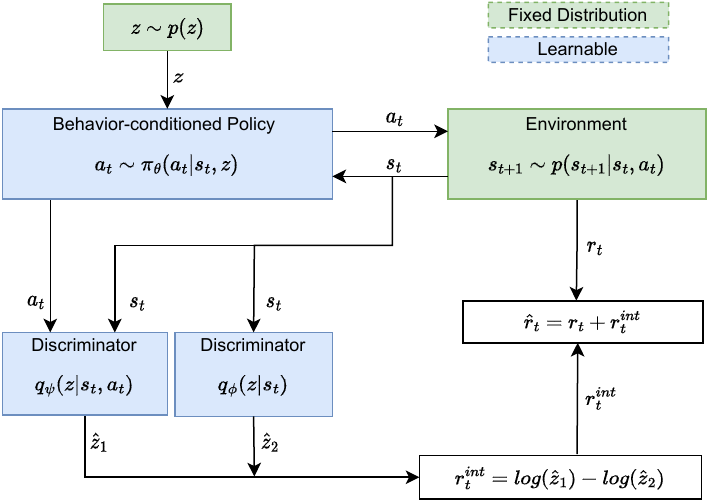}
    \caption{\textbf{Our framework for behavior-conditioned policy}. An intrinsic reward is computed based on discriminators $q_\psi$ and $q_\phi$, which encourages the diversity by indirectly maximizing the lower variation bound  $\mathbb{G}(\theta)$}
    \label{fig:diverse}
\end{figure}

Maximizing the intrinsic pseudo-reward through reinforcement learning allows maximizing $\mathbb{G}(\theta)$ despite sampling $s \sim \pi(z)$ from a non-differentiable simulator. In Eqn. \eqref{eq:2}, $a_t$ is sampled from a policy conditioned on behavior rather than a default policy, as generating diverse actions requires knowledge about $z$. The proposed intrinsic reward promotes action diversity while learning the main task.
\textcolor{black}{Overall, our intrinsic reward shares some similarity to DIAYN \cite{eysenbach2018diversity} in which both use token-conditioned policies. However, the difference is that our method encourages agents with different tokens to generate diverse actions for a given state instead visiting diverse states.}
Figure~\ref{fig:diverse} shows an overview of the interaction between the behavior-conditioned policy and the discriminators. 

\setlength{\textfloatsep}{1pt}
\begin{algorithm}[t]
\small
\caption{Behavior-conditioned policy for $N$ agents}
    \begin{algorithmic}[1]
        \State Initialize policy network $\pi_\theta$, value function $V_{\phi}$, discriminators \textcolor{blue}{$q_{\psi_{sa}}$} and \textcolor{blue}{$q_{\psi_{s}}$} \label{algol:4}
        \State Require: hyper-parameters $\textcolor{blue}{\alpha}, \gamma, \lambda, \epsilon, \textcolor{blue}{M}$
        \While {not converged}
            \State // Collect data in parallel             
            \For{$i=1,2,\dots,\text{N}$}
                \State \textcolor{blue}{Sample behavior $ z^i \sim p_M(z)$ }\label{algol:1}
                \State Sample \textcolor{blue}{behavior-conditioned policy $\pi_\theta ( \cdot, s_t | z )$} for $T_i$ timesteps, collecting \{$s_{t+1}^i, a_t^i, r_t^i$\} where $t \in [0, T_i]$
                 \label{algol:2}
                  \State \textcolor{blue}{ Modify reward by adding bonus intrinsic reward $r_t^i \leftarrow r_t^i + \alpha\left\{ log [q_{\psi_{sa}}(z \mid (s_t^i,a_t^i)]) - log[q_{\psi_s}(z \mid s_t^i)] \right\}$}
                \label{algol:3}                           
                \State Compute advantages using GAE \cite{schulman2015high} $\hat{A}_t^i= \sum_{l=0}^{T_i}(\gamma \lambda)^l(r_t^i+\gamma V_{\phi}\left(s_i^{t+1}\right)-V_{\phi}\left(s_t^i\right))$ \label{algol:5}
                
            \EndFor
            \State $\pi_{old} \leftarrow \pi_\theta$
            \State // Update Policy, Value Functions and Discriminators
            \For{$j=1 \text{ to } {epoch}_\pi$}
                \State Compute Ratio $k_t= \frac{\pi_\theta\left(a_t^i \mid o_t^i\right)}{\pi_{\text {old }}\left(a_t^i \mid o_t^i\right)}$
                \State $\mathbb{L}^{PPO\_clip}(\theta)=\sum_{t=1}^{T_{\max }} min\left( k_t \hat{A}_t^i, \text{clip}\left(k_t, 1-\epsilon, 1+\epsilon \right)  \hat{A}_t^i\right)$        
                \State Update $\theta$ using Adam w.r.t. $\mathbb{L}^{PPO\_clip}(\theta)$               
            \EndFor
            \For{$j=1 \text{ to epoch\_v} $}
                \State $\mathbb{L}^{V}(\phi)=-\sum_{i=1}^{N} \sum_{t=1}^{T_i}\left(\sum_{t^{\prime}>t} \gamma^{t^{\prime}-t} r^i_{t^{\prime}}-\right. \left.V_{\phi}\left(s_t^i\right)\right)^2$ \label{algol:4a}
                \State Update $\phi$ using Adam w.r.t. $\mathbb{L}^{V}(\phi)$ 
            \EndFor
            \For{$j=1 \text{ to epoch\_d} $}
                \State \textcolor{blue}{$\mathbb{L}^{D}(\psi_{sa})=-\sum_{i=1}^{N} \sum_{t=1}^{T_i}
                \left( z^i \cdot \log \left(q_{\psi_{sa}}(s_t^i, a_t^i)\right) \right) $}
                \State \textcolor{blue}{$\mathbb{L}^{D}(\psi_{s})=-\sum_{i=1}^{N} \sum_{t=1}^{T_i}
                \left( z^i \cdot \log \left(q_{\psi_{s}}(s_t^i)\right) \right) $}
                \State \textcolor{blue}{Update $\psi_{sa}$, $\psi_{s}$ using Adam w.r.t. $\mathbb{L}^{D}(\psi_{sa})$, $\mathbb{L}^{D}(\psi_{s})$ }\label{algol:6}
            \EndFor


        \EndWhile
    \end{algorithmic}
    \label{algo:1}
\end{algorithm}

\subsection{Training Procedure}
\label{sec:training}

We adapt the training procedure from \cite{long2018towards}, alternating between sampling trajectories and updating the policy via the PPO algorithm \cite{schulman2017proximal}. Each agent uses an identical policy to collect data until a batch is gathered.
Algorithm \ref{algo:1} outlines the training details. Key differences from \cite{long2018towards} are highlighted in \textcolor{blue}{blue}:
(1) At each episode start, agent $i$ samples a new behavior token $z^i \sim p_M(z)$, with $p_M(z)$ being a discrete uniform distribution with $M$ behaviors (Line \ref{algol:1}). \textcolor{black}{This token is mapped to a 32-dimensional continuous embedding.}
(2) Agents sample from a policy conditioned on $z^i$ (Line \ref{algol:2}), allowing for varied actions based on behavior.
(3) Intrinsic rewards for each agent are computed using discriminators $q_{\psi_{sa}}$ and $q_{\psi_{s}}$, parameterized by $\psi_{sa}$ and $\psi_{s}$ respectively, based on Eqn. \eqref{eq:2}. These rewards are added to the task reward in the replay buffer (Line \ref{algol:3}).
(4) We optimize discriminators $q_{\psi_{sa}}$ and $q_{\psi_{s}}$ with cross-entropy loss (Line \ref{algol:6}) using the Adam optimizer \cite{kingma2014adam}. \textcolor{black}{Adding one standard deviation of Gaussian noise to discriminator inputs helps prevent overfitting.} The loss is computed between predicted behavior tokens $\hat{z}$ from on-policy samples and ground truth behavior.

\section{Experiments}

We conduct comprehensive experiments to demonstrate the effectiveness of our method over previous solutions.
We simulate these experiments with new crowd behaviors not encountered during agent training.

\subsection{Experimental Setup}
\label{sec:experiment-setup}

\noindent\textbf{Implementation.}
We simulate a large-scale group of robots using Stage \cite{vaughan2008massively}, a popular robot simulator widely used in multi-agent research. Each agent is initialized as a non-holonomic differential drive robot \textcolor{black}{($0.5m \times 0.5m$)} equipped with a 2D-laser scanner to sense its surroundings. The 2D laser is set to 360 degrees FOV with a max range of 10m.

\begin{wraptable}{r}{0.5\linewidth}
\centering
\vspace{-2pt}
\resizebox{\linewidth}{!}{ 
\begin{tabular}{l|l}
\Xhline{1.5pt}
\textbf{Hyper-parameter} & \textbf{Value} \\ 
\Xhline{1.5pt}
Discount Factor $\gamma$ & 0.99 \\
PPO Smoothing $\lambda$ & 0.95 \\
PPO Clip Value $\epsilon$ & 0.1 \\
\# Epoch for Policy Network & 3 \\
\# Epoch For Value Network & 3 \\
\# Epoch for Discriminators & 1 \\
Advantange Weight $\alpha$ & \textcolor{black}{0.1} \\
PPO Learning Rates & 0.00005 \\
Discriminators Learning Rates & \textcolor{black}{0.00005} \\
\textcolor{black}{$\text{Epoch}_{\pi}$, $\text{Epoch}_d$, $\text{Epoch}_v$} & \textcolor{black}{3}  \\
\Xhline{1.5pt}
\end{tabular}}
\caption{\textbf{Hyper-parameters in our implementation.}}
\vspace{-15pt}
\label{table:0}
\end{wraptable}

For agent states, rewards and neural network architecture, we follow the same setup as \cite{long2018towards}. 
We make one change to the NN backbone by adding a behavior embedding derived from the behavior token, $z$, for the neural network input. 
Each discriminator is modeled with a two-layer feed-forward network with 128 hidden units and ReLU activations \cite{nair2010rectified}. 
Table \ref{table:0} lists the hyper-parameter settings. 

\noindent\textbf{Training Setup.}

We train agents in a realistic, heavily trafficked $20m \times 20m$ room. Random goals are placed at least $10m$ away from the agents' initial positions. A crash event is registered if laser-scan values fall below $0.5m$.

\begin{figure}[t]
    \centering
    \begin{minipage}{\columnwidth}
    \includegraphics[width=0.49\linewidth]{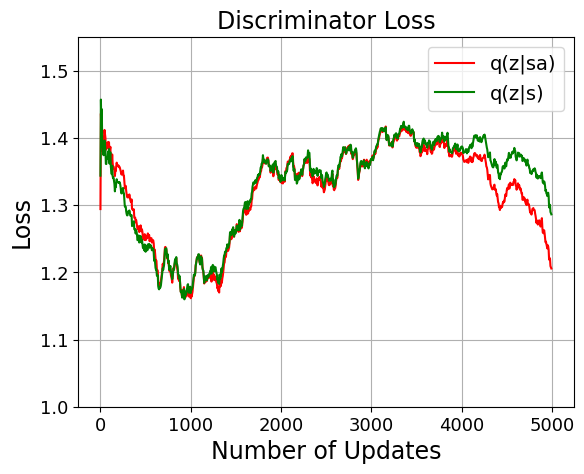}
    \hfill
    \includegraphics[width=0.49\linewidth]{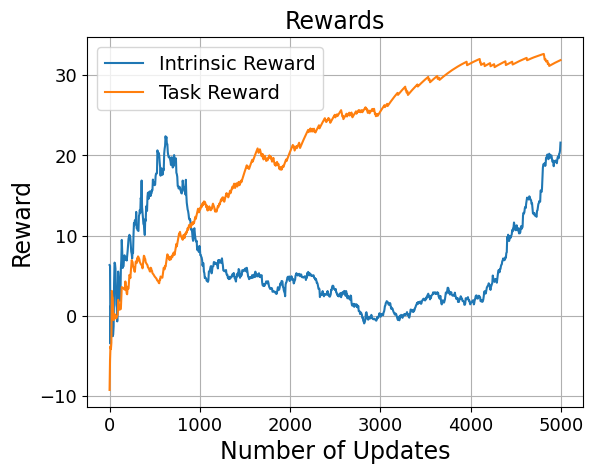}        
    \end{minipage}
    \caption{\textcolor{black}{The discriminator loss and reward curves.}}
    \label{fig:loss}     
\end{figure}

\noindent\textbf{Testing Setup.} 
\textcolor{black} {To evaluate the robustness of our policy in unseen crowd behaviors, we propose a novel set of testing conditions with the criteria that the movements of the other agents must be unique and not encountered during training. Also, we omit static obstacles as we focus on dynamic obstacle avoidance. In these environments, our agent interacts with other agents which exhibit dynamic movements beyond our control}. For clarity, we refer to other agents as pedestrians for the remainder of this section. In total, we design \textcolor{black}{six} different pedestrian setups to be evaluated for each study:
\begin{enumerate}
\item \textit{Non-homogeneous} (NH): To achieve non-homogeneous behaviors, each pedestrian utilizes a distinct policy instance, initialized with a unique seed, to generate varied experiences during training.
\item \textit{Invisible} (IN): Similar to 1), but our robot is invisible to pedestrians achieved by lowering our agent's height below the pedestrians' sensors. This reflects the non-reactive individuals in the real world. 
\item \textit{Variability} (VA): Similar to 2), pedestrians are invisible but receive random speed multipliers (0.5-1.5) at each episode start, representing real-world walking speed variability.
\item \textit{Sub-optimal} (SO): Similar to 1), but the policy is trained for half the period (2.5k updates instead of 5k), simulating sub-optimal walking trajectories.
\item \textit{Velocity-obstacle} (VO): We utilizes ground truth positions of all pedestrians to compute permissible velocities, following the method in \cite{van2011reciprocal}.
\item \textit{Social force} (SF): We use a force-based system to anticipate pedestrian movements following \cite{helbing1995social}, and utilize ground positions for prediction similar to VO.

\end{enumerate}

We consider a mixture of learning-based (NH,IN,VA,SO) and non-learning based (VO,SF) policies. Four of them (IN,SO,VO,SF) are challenging with the non-reactive pedestrians. All evaluations are repeated for 1000 episodes. If not stated explicitly, we set the number of robots $N=5$ (1 agent and 4 pedestrians) and number of behaviors $M=5$. Agents, pedestrians and goals are spawned similarly to training. During testing, each agent utilises a fixed behavior token, $z=0$ for all episodes.
Our primary metric is the `success rate', without further classifying the non-successful episodes, as collisions are the primary reason instead of timeouts.

\begin{table}[t]
\centering
\begin{adjustbox}{width=\linewidth,center}
\color{black}\begin{tabular}{c|c|c|cccccc|c}
\Xhline{1.5pt}
\multirow{2}{*}{\textbf{M}} & \multirow{2}{*}{\textbf{\#Updates}}  & \multirow{2}{*}{\textbf{\#Updates/M}} & \multicolumn{6}{c|}{\textbf{Pedestrian Type}} & \multirow{2}{*}{\textbf{Avg}} \\ \cline{4-9}
& & & \multicolumn{1}{l|} {\textbf{NH}} & \multicolumn{1}{l|}{\textbf{IN}} & \multicolumn{1}{l|}{\textbf{VA}} & \multicolumn{1}{l|}{\textbf{SO}} & \multicolumn{1}{l|}{\textbf{VO}} & \textbf{SF} \\ \Xhline{1.5pt}
1 &5K & 5K & \multicolumn{1}{c}{0.63} & \multicolumn{1}{c}{0.55} & \multicolumn{1}{c}{0.59}& \multicolumn{1}{c}{0.52}& \multicolumn{1}{c}{0.59} & \multicolumn{1}{c|}{0.43} & 0.55 \\
5 &5K & 1K & \multicolumn{1}{c}{\textbf{0.87}} & \multicolumn{1}{c}{\textbf{0.85}} & \multicolumn{1}{c}{\textbf{0.86}}& \multicolumn{1}{c}{\textbf{0.72}}& \multicolumn{1}{c}{\textbf{0.78}} & \multicolumn{1}{c|}{\textbf{0.77}} & \textbf{0.81} \\
10 &5K & 500 & \multicolumn{1}{c}{0.80} & \multicolumn{1}{c}{0.75} & \multicolumn{1}{c}{0.82}& \multicolumn{1}{c}{0.72}& \multicolumn{1}{c}{0.76} & \multicolumn{1}{c|}{0.69} & 0.76 \\
20 &5K & 250 & \multicolumn{1}{c}{0.61} & \multicolumn{1}{c}{0.63} & \multicolumn{1}{c}{0.54}& \multicolumn{1}{c}{0.59}& \multicolumn{1}{c}{0.59} & \multicolumn{1}{c|}{0.51} & 0.58 \\
\Xhline{1pt}
10 &10K & 1K & \multicolumn{1}{c}{0.88} & \multicolumn{1}{c}{0.86} & \multicolumn{1}{c}{0.86}& \multicolumn{1}{c}{0.76}& \multicolumn{1}{c}{0.77} & \multicolumn{1}{c|}{0.78} & 0.82 \\
20 &20K & 1K & \multicolumn{1}{c}{0.89} & \multicolumn{1}{c}{0.87} & \multicolumn{1}{c}{0.85}& \multicolumn{1}{c}{0.77}& \multicolumn{1}{c}{0.78} & \multicolumn{1}{c|}{0.79} & 0.82 \\
\Xhline{1.5pt}

\end{tabular}
\end{adjustbox}
\caption{\textbf{Impact of the number of behaviors $M$}. \textcolor{black}{Policies are evaluated under six diverse unseen pedestrian setups.}}
\label{table:1}
\end{table}

\subsection{Training Results}
\label{sec:evaluate-training}

Figure \ref{fig:loss} shows the stability of the discriminators and the intrinsic reward during training. 
The discriminator loss and intrinsic reward steadily improves until $\sim700$ updates, which then flattens until $\sim4000$ updates, possibly due to novel state-action exploration. Subsequently, both curves continue to improve again until the end of training where the main task has converged. The best advantage weight $\alpha$, which combines the task and intrinsic reward is 0.1.

\begin{wrapfigure}{r}{0.55\linewidth}
    \vspace{-15pt}
    \centering
    \includegraphics[width=\linewidth]{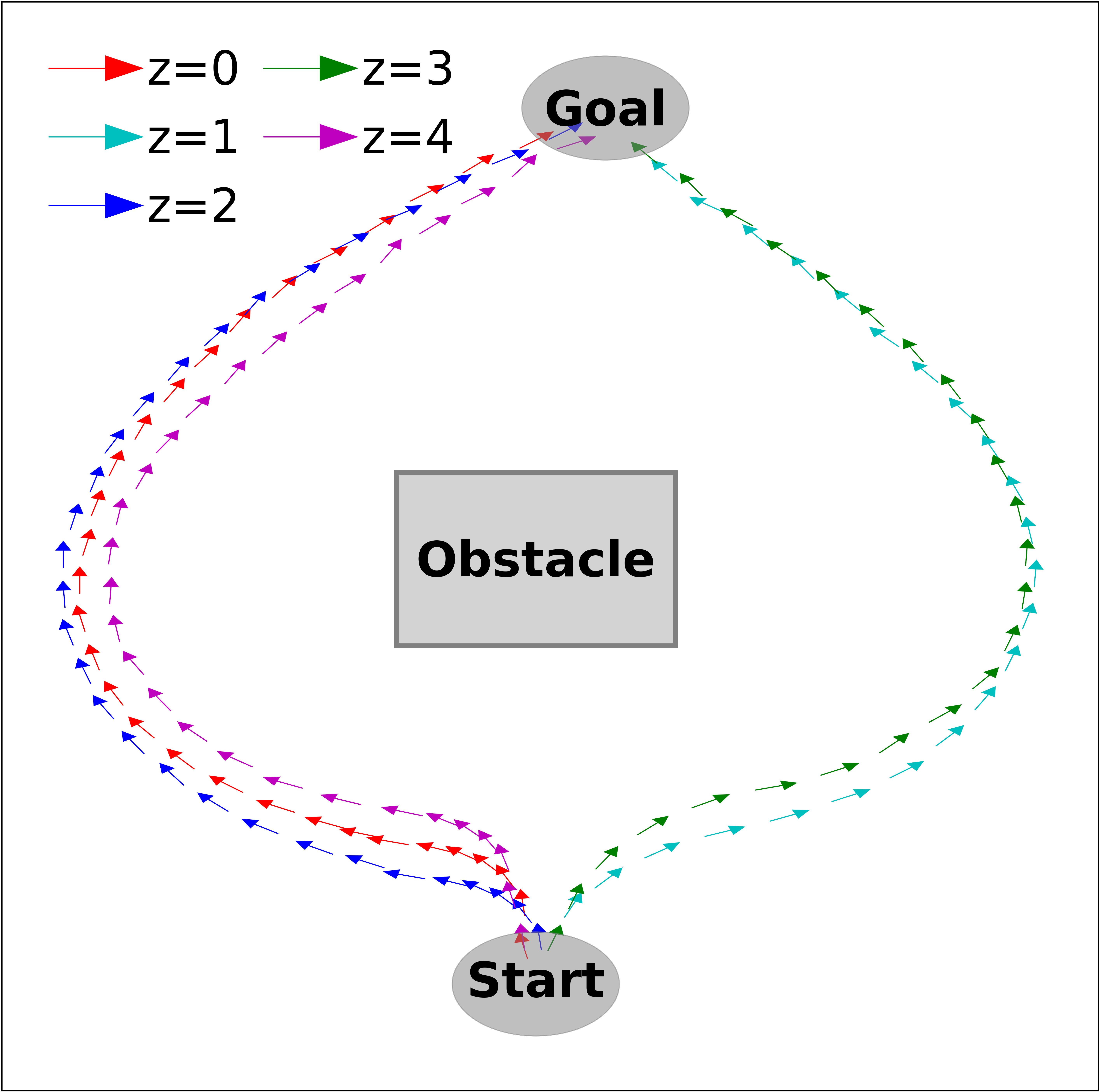}
    \caption{Agents' varied paths based on sampled behaviors.}
    \label{fig:qual}
    \vspace{-10pt}
\end{wrapfigure}

Next, we investigate if behavior-conditioned policies exhibit different behaviors to reach the desired goal.
We record the behaviors of agents starting from the origin and reaching a fixed goal behind a static obstacle. The trajectories of the 5 agents can be seen in Figure \ref{fig:qual}. These agents exhibit different behaviors when conditioned on different tokens $z$. While passing the obstacle, some agents are left-inclined whereas some are right-inclined which is consistent with the motivation illustrated in Figure \ref{fig:high}. 
Additionally, we also observe the velocity diversity which is represented by the length of the arrow. 
For example, the blue agent ($z=2$) travels slightly faster than the purple agent ($z=4$) when taking a slightly longer route. 
While the diversity may appear slight, deploying it with multiple agents generates vastly different training trajectories, enhancing agent robustness.
More qualitative examples about the diversity among dynamic obstacles are available on \href{https://youtu.be/EevMn2-ZNng}{https://youtu.be/EevMn2-ZNng}.

\subsection{Impact of the Number of Behaviors}
\label{subs:m-behavior}

We assess diversity's impact on agent robustness and its scalability with behavior count $M$.
When $M=1$, all agents share the same behavior, equivalent to the baseline policy in \cite{long2018towards}.
Our results are presented in Table \ref{table:1}. 
We have the following three observations. \textcolor{black}{First, having non-reactive pedestrians (NH,IN,VA,SO) is generally a more difficult task with fewer successful runs}. Second, adding diversity ($M\ne 1$) outperforms the default policy ($M=1$) for all pedestrian types, including the challenging non-reactive pedestrians. This validates the effectiveness of our proposed method. Third, the optimal number of behaviors is $M=5$ and the effect of diversity starts to diminish as we scale to higher values of $M=10$ and $20$. 
\textcolor{black}{We hypothesize the diminishing effect is resulted from less frequent sampling when $M$ increases. 
To investigate this, we increase ${num\_updates}/M$ for $M=10$ and $20$ to match $M=5$ and find that the performance of $M=10$ and $M=20$ could match that of $M=5$, validating our hypothesis.
However, this comes at an expense of more updates. Overall, $M=5$ provides a good balance between creating good diversity and sample efficiency.}

\subsection{Scalability}
\label{subs:n-agents}
Here, we investigate the effect of increased numbers of agents $N$ and report the results in Table \ref{table:2}. 
As the number of agents, $N$, increases to 10 and 20, the scenes become more crowded, making it harder for them to reach their goals. This negatively affects the convergence speed, as agents get restarted more frequently due to collisions or other obstacles. Despite this, by adding diversity to the policy, we achieve consistent performance improvements across all pedestrian cases.

\begin{table}[t]
\centering
\begin{adjustbox}{width=\linewidth,center}
\color{black}\begin{tabular}{c|c|cccccc|c}
\Xhline{1.5pt}
\multirow{2}{*}{\textbf{$N$}}  & \multirow{2}{*}{\textbf{Diversity}} & \multicolumn{6}{c|}{\textbf{Pedestrian Type}} & \multirow{2}{*}{\textbf{Avg}} \\ \cline{3-8}
&   & \multicolumn{1}{l|} {\textbf{NH}} & \multicolumn{1}{l|}{\textbf{IN}} & \multicolumn{1}{l|}{\textbf{VA}} & \multicolumn{1}{l|}{\textbf{SO}} & \multicolumn{1}{l|}{\textbf{VO}} & \textbf{SF} \\ \Xhline{1.5pt}

5  & No & \multicolumn{1}{c}{0.63} & \multicolumn{1}{c}{0.55} & \multicolumn{1}{c}{0.59}& \multicolumn{1}{c}{0.52}& \multicolumn{1}{c}{0.59} & \multicolumn{1}{c|}{0.43} & 0.55 \\
5  & Yes & \multicolumn{1}{c}{\textbf{0.87}} & \multicolumn{1}{c}{\textbf{0.85}} & \multicolumn{1}{c}{\textbf{0.86}}& \multicolumn{1}{c}{\textbf{0.72}}& \multicolumn{1}{c}{\textbf{0.78}} & \multicolumn{1}{c|}{\textbf{0.77}} & \textbf{0.81} \\
\Xhline{0.5pt}                  
10  & No & \multicolumn{1}{c}{0.41} & \multicolumn{1}{c}{0.38} & \multicolumn{1}{c}{0.42}& \multicolumn{1}{c}{0.37}& \multicolumn{1}{c}{0.31} & \multicolumn{1}{c|}{0.28} & 0.35 \\
10  & Yes & \multicolumn{1}{c}{0.83} & \multicolumn{1}{c}{0.78} & \multicolumn{1}{c}{0.77}& \multicolumn{1}{c}{0.65}& \multicolumn{1}{c}{0.75} & \multicolumn{1}{c|}{0.52} & 0.72 \\
\Xhline{0.5pt}
20  & No & \multicolumn{1}{c}{0.47} & \multicolumn{1}{c}{0.43} & \multicolumn{1}{c}{0.42}& \multicolumn{1}{c}{0.39}& \multicolumn{1}{c}{0.36} & \multicolumn{1}{c|}{0.32} & 0.38 \\
20  & Yes & \multicolumn{1}{c}{0.58} & \multicolumn{1}{c}{0.50} & \multicolumn{1}{c}{0.58}& \multicolumn{1}{c}{0.47}& \multicolumn{1}{c}{0.42} & \multicolumn{1}{c|}{0.49} & 0.50 \\
\Xhline{1.5pt}
\end{tabular}
\end{adjustbox}
\caption{\textbf{Impact of the number of agents $N$.}}
\label{table:2}
\end{table}

\begin{table}[t]
\centering
\begin{adjustbox}{width=\linewidth,center}
\color{black}\begin{tabular}{c|c|cccccc|c|c}
\Xhline{1.5pt}
\multirow{2}{*}{\textbf{Intrinsic Reward}} & \multirow{2}{*}{\textbf{$\alpha_\text{ best}$}} & \multicolumn{6}{c|}{\textbf{Pedestrian Type}} & \multirow{2}{*}{\textbf{Avg}} & \multirow{2}{*}{\textbf{$\mathcal{D}$}} \\ \cline{3-8}
& & \multicolumn{1}{l|} {\textbf{NH}} & \multicolumn{1}{l|}{\textbf{IN}} & \multicolumn{1}{l|}{\textbf{VA}} & \multicolumn{1}{l|}{\textbf{SO}} & \multicolumn{1}{l|}{\textbf{VO}} & \textbf{SF} & \\ \Xhline{1.5pt}
$log[q_1(z\mid s,a)] - log[q_2(z\mid s)]$ & 0.10 & \multicolumn{1}{c}{\textbf{0.87}} & \multicolumn{1}{c}{\textbf{0.85}} & \multicolumn{1}{c}{\textbf{0.86}}& \multicolumn{1}{c}{\textbf{0.72}}& \multicolumn{1}{c}{\textbf{0.78}} & \multicolumn{1}{c|}{\textbf{0.77}} & \textbf{0.81} & 1.10 \\
$log[q(z\mid s,a)]$ & 0.01 & \multicolumn{1}{c}{0.88} & \multicolumn{1}{c}{0.77} & \multicolumn{1}{c}{0.86}& \multicolumn{1}{c}{0.70}& \multicolumn{1}{c}{0.75} & \multicolumn{1}{c|}{0.72} & 0.78 & 0.48 \\
$log[q(z\mid s)]$ \text{\cite{eysenbach2018diversity}} & 0.03 &\multicolumn{1}{c}{0.75} & \multicolumn{1}{c}{0.75} & \multicolumn{1}{c}{0.69}& \multicolumn{1}{c}{0.71}& \multicolumn{1}{c}{0.73} & \multicolumn{1}{c|}{0.58} & 0.70 & 0.17 \\
None & 0 & \multicolumn{1}{c}{0.63}  & \multicolumn{1}{c}{0.55} & \multicolumn{1}{c}{0.59}& \multicolumn{1}{c}{0.52}& \multicolumn{1}{c}{0.59} & \multicolumn{1}{c|}{0.43} & 0.55 & 0 \\
\Xhline{1.5pt}
\end{tabular}
\end{adjustbox}
\caption{\textbf{Policies trained using different intrinsic rewards}.  \textcolor{black}{$\mathcal{D}$ is a measure of action diversity between agents}}  
\label{table:3}
\end{table}

\begin{table*}[t]
\centering
\begin{adjustbox}{width=0.75\textwidth,center}
\color{black}\begin{tabular}{c|c|c|c|c|c|c|c} 
\Xhline{1.5pt}
\multirow{2}{*}{\textbf{Metrics}} & \multirow{2}{*}{\textbf{Policy}} & \multicolumn{4}{c}{\textbf{Pedestrian Type}} \\ \cline{3-8}
 & & \textbf{NH} & \textbf{IN} & \textbf{VA} & \textbf{SO} & \textbf{VO} & \textbf{SF} \\ \Xhline{1.5pt}
 
\multirow{3}{*}{Success Rate $\uparrow$} 
& Basic & \multicolumn{1}{c|}{0.63}  & \multicolumn{1}{c|}{0.55} & \multicolumn{1}{c|}{0.59}& \multicolumn{1}{c|}{0.52}& \multicolumn{1}{c|}{0.59} & \multicolumn{1}{c}{0.43} \\ 
& Safe & \multicolumn{1}{c|}{0.86} & \multicolumn{1}{c|}{0.77} & \multicolumn{1}{c|}{0.81} & 0.67 & 0.69 & 0.61 \\ 
& Ours & \multicolumn{1}{c|}{\textbf{0.87}} & \multicolumn{1}{c|}{\textbf{0.85}} & \multicolumn{1}{c|}{\textbf{0.86}}& \multicolumn{1}{c|}{\textbf{0.72}}& \multicolumn{1}{c|}{\textbf{0.78}} & \multicolumn{1}{c}{\textbf{0.77}} \\ \hline
 
\multirow{3}{*}{Extra Time (s) $\downarrow$} 
& Basic & \multicolumn{1}{c|}{2.833 $\pm$ 2.439} & \multicolumn{1}{c|}{3.366 $\pm$ 2.621} & \multicolumn{1}{c|}{3.724 $\pm$ 2.219} & 2.511 $\pm$ 1.751 & 3.158 $\pm$ 2.121 & 2.997 $\pm$ 1.136 \\ 
& Safe & \multicolumn{1}{c|}{5.041 $\pm$ 2.356} & \multicolumn{1}{c|}{5.102 $\pm$ 2.719} & \multicolumn{1}{c|}{5.248 $\pm$ 2.658} & 5.100 $\pm$ 2.335 & 5.217 $\pm$ 2.454 & 4.813 $\pm$ 2.348 \\ 
& Ours & \multicolumn{1}{c|}{\textbf{2.712 $\pm$ 2.259}} & \multicolumn{1}{c|}{\textbf{2.902 $\pm$ 2.671}} & \multicolumn{1}{c|}{\textbf{2.714 $\pm$ 2.427}} & \textbf{2.336 $\pm$ 0.995} & \textbf{2.119 $\pm$ 1.038} & \textbf{2.202 $\pm$ 1.344
} \\ \hline
 
\multirow{3}{*}{Extra Distance (m) $\downarrow$} 
& Basic & \multicolumn{1}{c|}{4.811 $\pm$ 4.011} & \multicolumn{1}{c|}{4.123 $\pm$ 5.637} & \multicolumn{1}{c|}{5.717 $\pm$ 3.013} & 4.197 $\pm$ 5.873  & 5.887 $\pm$ 4.187  & 3.321 $\pm$ 3.899 \\ 
& Safe & \multicolumn{1}{c|}{10.099 $\pm$ 4.452} & \multicolumn{1}{c|}{10.734 $\pm$ 5.177} & \multicolumn{1}{c|}{16.601 $\pm$ 5.235} & 10.116 $\pm$ 4.182  & 9.870 $\pm$ 3.946  & 10.024 $\pm$ 4.275 \\ 
& Ours & \multicolumn{1}{c|}{\textbf{3.667 $\pm$ 3.587}} & \multicolumn{1}{c|}{\textbf{3.930 $\pm$ 4.024}} & \multicolumn{1}{c|}{\textbf{4.262 $\pm$ 4.489}} & \textbf{3.217 $\pm$ 2.309}  & \textbf{2.492 $\pm$ 1.719}   & \textbf{2.662 $\pm$ 2.112} \\ \hline
 
\multirow{3}{*}{Average Speed (m/s) $\uparrow$} 
& Basic & \multicolumn{1}{c|}{0.919 $\pm$ 0.096} & \multicolumn{1}{c|}{0.927 $\pm$ 0.088} & \multicolumn{1}{c|}{0.917 $\pm$ 0.87} & 0.922 $\pm$ 0.068 & 0.920 $\pm$ 0.079  & 0.910 $\pm$ 0.087 \\ 
& Safe & \multicolumn{1}{c|}{0.810 $\pm$ 0.091} & \multicolumn{1}{c|}{0.795 $\pm$ 0.098} & \multicolumn{1}{c|}{0.779 $\pm$ 0.100} & 0.811 $\pm$ 0.084 & 0.811 $\pm$ 0.084  & 0.786 $\pm$ 0.097 \\ 
& Ours & \multicolumn{1}{c|}{\textbf{0.957 $\pm$ 0.059}} & \multicolumn{1}{c|}{\textbf{0.955 $\pm$ 0.061}} & \multicolumn{1}{c|}{\textbf{0.942 $\pm$ 0.076}} & \textbf{0.960 $\pm$ 0.057} & \textbf{0.966 $\pm$ 0.039} & \textbf{0.965 $\pm$ 0.043} \\ \Xhline{1.5pt}
\end{tabular}
\end{adjustbox}
\caption{\textbf{Comparisons with baseline methods using different metrics averaged across 1000 episodes.}}
\vspace{-10pt}
\label{table:4}
\end{table*}

\subsection{\textcolor{black}{Intrinsic Rewards}}
\label{subs:ablation}
In Section \ref{sec:behavior-formulation}, we formulate a cost function to promote diversity among agents. In the formulation, we require that diverse actions performed for different $z$ should arise for every state instead of exploiting only certain states. This is achieved using a regularizer as part of the intrinsic reward proposed in Eqn. \eqref{eq:2}. 
From ablation experiments reported in Table \ref{table:3}, we observe that the policy trained with the intrinsic reward containing the regularization term, $- log(q\mid s)$, outperforms the policy without this term. The performance improvement is consistent in all pedestrian setups including the challenging non-reactive pedestrians. Despite this, the policy trained without regularization still outperforms the base policy without the intrinsic reward.

Next, we compare our method with state-space exploration based intrinsic rewards, DIYAN, from \cite{eysenbach2018diversity} which may implicitly encourage action diversity through novel state exploration. However, it still lacks action diversity compared to our proposed intrinsic reward in Eqn.\eqref{eq:2}, where the diversity of the action is explicitly encouraged. To measure the action diversity, we also introduce a new metric, $\mathcal{D}$, using the KL divergence of action distributions between pairwise agents:
$$
\mathcal{D} =  \frac{1}{|\tau| N_{i\ne j}} \sum_{s\in \tau } \sum_{i\ne j}{\text{KL}\left(\pi(a|s,z=i) \| \pi(a|s,z=j) \right)}
$$
\textcolor{black}{where $\tau$ denotes a trajectory. Specifically, we collect a trajectory of 1000 steps using the trained policy with no intrinsic reward. From Table \ref{table:3}, our proposed method achieves higher action diversity $\mathcal{D}$ than the state-space exploration based intrinsic rewards. Also, we observe that higher values of $\mathcal{D}$ get translated into higher robustness in unseen crowd behaviors, achieving a greater success rate.}\\

\subsection{Comparisons with Prior Work}
\label{subs:trade}

We quantitatively \textcolor{black}{compare} our proposed behavior-conditioned policy with existing solutions to demonstrate its robustness. 
In particular, we set up the baseline method as described in \cite{long2018towards}, equivalent to our proposed method with $M=1$. Additionally, we added a safe policy proposed in \cite{jin2020mapless}, which uses safety zone rewards to encourage safe behaviors, which could crash less in unseen crowd movements. 
For our proposed method, we utilize the model trained with $M=5$ and $N=5$. 
Table \ref{table:4} shows the comparison results against different metrics across 1000 episodes. Each metrics (success rate, extra time, extra distance, average speed) are similarly defined like in \cite{long2018towards}.

Our proposed method consistently outperforms others across various pedestrian types, suggesting robust strategies for handling diverse crowd behaviors effectively. While the safe policy achieves a higher success rate than the base policy, it slightly falls short of our proposed policy. Collisions primarily contribute to non-successful episodes, surpassing timeouts.
The safe policy with a safety buffer performs well in reactive setups (NH, VA), closely matching our policy's results. However, it struggles in non-reactive setups (IN, SO, VO, SF). The conservative behavior of the safe policy reduces collisions but increases time and distance compared to our proposed policy, sacrificing other metrics. Specifically, both time and distance taken by the safe policy are more than double those of our proposed policy.

\begin{figure}[]
    \centering
    \begin{minipage}{\columnwidth}
    \includegraphics[width=0.49\linewidth]{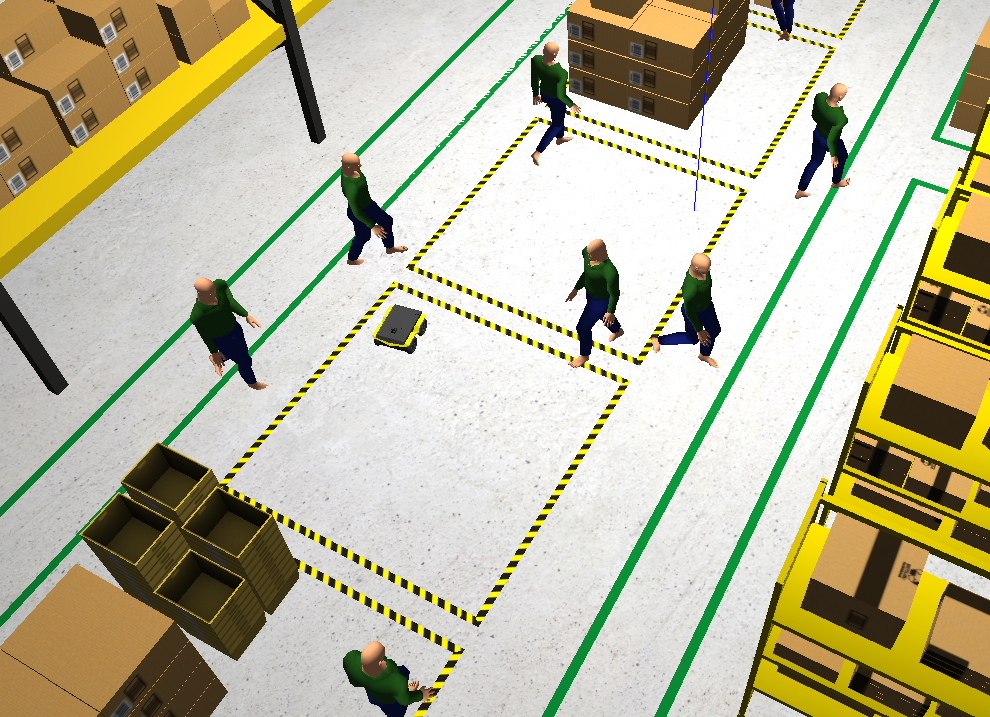}
    \hfill
    \includegraphics[width=0.49\linewidth]{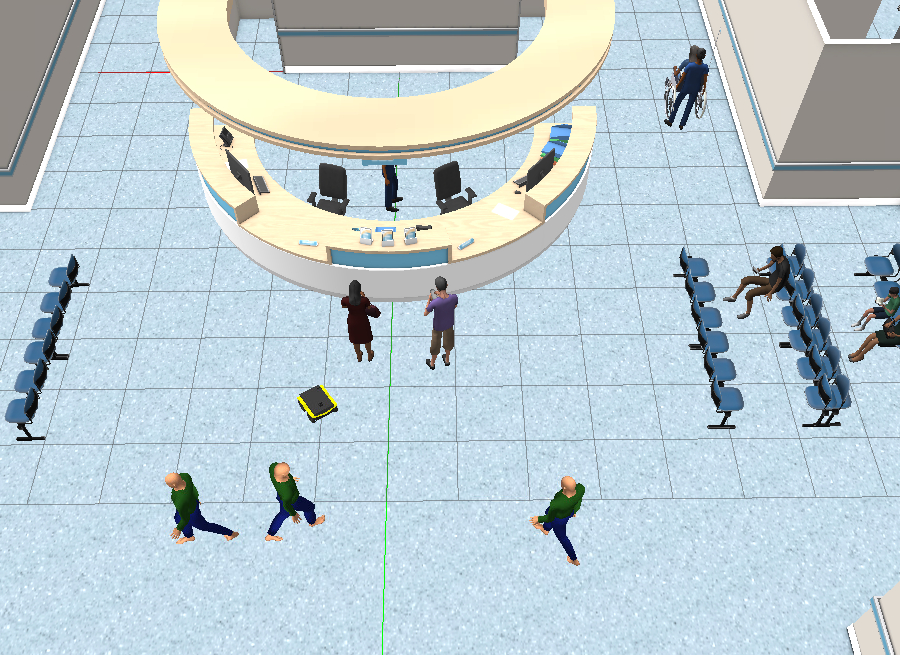}        
    \end{minipage}
    \caption{\textbf{Testing our method in Gazebo with more realistic scenarios}. Map settings: (Left) Warehouse (Right) Hospital}
    \label{fig:gazebo}     
\end{figure}

\subsection{Realistic Deployment}
To validate our method in a more realistic setup, we deploy our best performing policy ($N=5$, $M=5$) on a Jackal robot in Gazebo simulator \cite{koenig2004design}. We utilize two maps (warehouse and hospital) from Arena-ROSNAV-3D \cite{arena-bench} as seen in Figure. \ref{fig:gazebo}. 
For each episode, we randomly assign the start and goal position with 3 to 8 pedestrians within an open area of each map $(\sim10m\times10m)$. The pedestrians movements are simulated using the social force model \cite{helbing1995social}.

\begin{wraptable}{}{0.5\linewidth}
\vspace{-10pt}
\centering
\resizebox{\linewidth}{!}{ 
\begin{tabular}{c|c|c}
\Xhline{1.5pt}
\textbf{Diversity} & \textbf{Warehouse} & \textbf{Hospital} \\
\Xhline{0.5pt}
No & 0.40 & 0.37 \\
Yes & \textbf{0.81} & \textbf{0.69} \\
\Xhline{1.5pt}
\end{tabular}}
\caption{\textbf{Success rate out of 100 episodes}}
\vspace{-10pt}
\label{table:gazebo}
\end{wraptable}
Table \ref{table:gazebo} shows the success rate from 100 episodes with and without the diversity consideration. 
Our method proves to be equally effective for realistic scenes, outperforming the baseline method when diversity is used during training. 
More qualitative examples for realistic scenes are available on \href{https://youtu.be/EevMn2-ZNng}{https://youtu.be/EevMn2-ZNng}.

\section{Conclusion}
This paper introduces a framework to increase an agent's ability to generalize to unseen crowd behaviors by utilizing diverse behaviors in a sample-efficient manner. 
Adding diversity in a multi-agent framework implicitly provides each agent with a more varied range of experiences, hence increasing its generalizability of unseen crowd behaviors. 
We demonstrate the robustness of the proposed method in an extensive set of evaluation scenes containing challenging pedestrians' behaviors. We also validate the scalability of our solution and practicality in realistic scenes. Our experiments also demonstrate that our method improves the success rates without negatively affecting other important metrics. 




\section*{ACKNOWLEDGMENT}
This study is supported under RIE2020 Industry Alignment Fund – Industry Collaboration Projects (IAF-ICP) Funding Initiative, as well as cash and in-kind contribution from the industry partner(s).

\bibliographystyle{IEEEtran}
\bibliography{IEEEexample}

\end{document}